\pdfoutput=1
\documentclass[11pt]{article}

\usepackage[]{acl}

\usepackage{times}
\usepackage{latexsym}
\usepackage[T1]{fontenc}
\usepackage[utf8]{inputenc}
\usepackage{microtype}
\usepackage{graphicx}
\usepackage{booktabs}
\usepackage{makecell}
\usepackage{tabularx}
\usepackage{mathtools}
\usepackage{multirow}

\title{Building Knowledge-Guided Lexica to Model Cultural Variation}

\author{Shreya Havaldar$^\dag$, Salvatore Giorgi$^\dag$, Sunny Rai$^\dag$, Young-Min Cho$^\dag$, \\ \bf Thomas Talhelm$^\diamond$, Sharath Chandra Guntuku$^\dag$, \& Lyle Ungar$^\dag$ \\
$^\dag$University of Pennsylvania, $^\diamond$University of Chicago \\
\texttt{\{shreyah,ungar\}@seas.upenn.edu} \\}

\begin{document}

\maketitle

\begin{abstract}
Cultural variation exists between nations (e.g., the United States vs. China), but also \textit{within} regions (e.g., California vs. Texas, Los Angeles vs. San Francisco). Measuring this regional cultural variation can illuminate how and why people think and behave differently. Historically, it has been difficult to computationally model cultural variation due to a lack of training data and scalability constraints.  In this work, we introduce a new research problem for the NLP community: \textit{How do we measure variation in cultural constructs across regions using language?} We then provide a scalable solution: building knowledge-guided lexica to model cultural variation, encouraging future work at the intersection of NLP and cultural understanding. We also highlight modern LLMs' failure to measure cultural variation or generate culturally varied language.

\end{abstract}

\section{Introduction}

People think and behave differently around the world. This is partly due to \textit{cultural variation}, or the differences among individuals that exist due to some form of social learning \cite{cohen2001cultural}. 

Having a computational method that utilizes language to measure cultural variation could help us better understand the way people communicate \cite{tsai2006cultural, oishi2009cross}, build more culturally-aware NLP systems \cite{hovy-yang-2021-importance}, and advance interdisciplinary research in anthropology, cultural psychology, etc. However, due to a lack of data and scalability constraints, few such methods exist. 

In this paper, we present \textit{measuring regional variation in culture} as a problem of interest for the NLP community. We highlight how large language models (LLMs) struggle with this task, and build a knowledge-guided lexical model as a scalable and reliable solution. Specifically, we focus on measuring the cultural dimenision of \textit{individualism and collectivism}\footnote{Cultural psychologists have quantified axes on which culture differs, also called \textit{cultural dimensions}. A key cultural dimension is called \textit{individualism vs. collectivism} \cite{hofstede2011dimensionalizing}. Collectivism stresses the importance of the community, while individualism focuses on each person's rights and concerns. This dimension has been shown to influence behaviors like voting, donating, etc. } across the United States (US) using geolocated Tweets.

\begin{figure}[t]
    \centering
    \includegraphics[width=\linewidth]{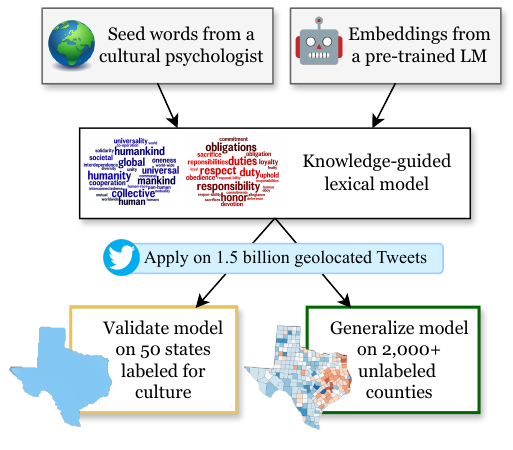}
    \caption{We build knowledge-guided lexica to model cultural variation. Our method encodes domain knowledge via seed words based on cultural psychology theory. We use embeddings to transform these seed words into a high-validity lexical model that successfully measures cultural variation across the US. }
    \label{fig:method}
\end{figure}

Historically, measuring cultural dimensions across regions has been mostly done through questionnaires, such as the World Values Survey (WVS) \cite{haerpfer2020world}. However, questionnaires are time-consuming and heavily restricted in scope; the most recent WVS wave required four years and averaged only 52 participants per US state. Recent work probes LLMs for cultural values \cite{arora-etal-2023-probing}, but these LLMs do not reflect all cultures equally \cite{havaldar-etal-2023-multilingual}. Therefore, relying on LLMs to measure culture is risky, as they may not generalize well to different populations.

The overhead of traditional survey-based approaches and inconsistent cultural awareness of existing LLMs motivate computational methods that rely on \textit{existing language data} to measure cultural variation instead. We specifically seek to use geolocated Twitter data --- instead of selecting a small portion of people to represent a state or county (like traditional survey-based methods), we instead use a massive amount of Tweets from that region, thus gaining a larger and more holistic representation of a region's population. 

We also seek to leverage \textit{domain expertise} from cultural psychologists. Cultural psychologists have spent decades developing non-computational tools to measure cultural constructs like individualism and collectivism \cite{talhelm2014large}. By encoding this domain knowledge via a set of expert-curated seed words, we can create a method to measure culture that is both scalable and grounded in cultural psychology theory.

In this paper, as an example of cultural variation, we will measure individualism and collectivism across US counties using the following resources: 
\begin{itemize}
\vspace{-0.1cm}
    \itemsep 0cm
    \item Domain knowledge from an expert psychologist researching collectivism.
    \item An open-source corpus (see Appendix~\ref{app:corpus}) of 1.5 billion geolocated Tweets from 6 million US users \cite{giorgi-etal-2018-remarkable}.
    \item Collectivism indicators (survey data, living arrangements, religiosity, and ingroup bias) to validate our results for fifty US states. \cite{vandello1999patterns, pelham2022truly}.
\end{itemize}

\paragraph{Challenges with deep learning approaches.} A modern NLP solution to measure culture today would take the form of either labeling data and training a model, or prompting a pre-trained LM. However, LLMs have been shown to lack cultural awareness \cite{havaldar-etal-2023-multilingual, liu2023multilingual}, so cultural insights from these models may be incorrect or untrustworthy. 

Additionally, classifying 1.5 billion Tweets requires a sizable amount of labeled training data, and training on such a large dataset is not computationally scalable. For instance, running our entire corpus through GPT-4 would cost roughly \$900,000 (see Appendix~\ref{app:calculations}). 

At a higher level, building a Tweet-level deep learning model to predict culture is impractical. Most of an individual's language does not indicate their cultural beliefs. Given this sparsity, labeling enough Tweets to train an adequate model is prohibitively expensive.

\vspace{0.25cm}
\noindent Our method builds upon a line of work in NLP called lexicon induction \cite{araque2020moralstrength,buechel2020learning,geng2022inducing, havaldar2023comparing}, which analyzes massive corpora in NLP without solely relying on deep learning. Past work mainly builds lexica for sentiment, emotion, etc. We uniquely focus on lexicon induction in the domain where little labeled data exists and not every utterance can be relevantly labeled.

\vspace{0.25cm}
\noindent Our contributions are as follows:
\begin{enumerate}
    \vspace{-0.2cm}
    \itemsep 0cm
    \item We present \textit{measuring regional variation in culture} as a problem of interest for the NLP community.
    \item We develop knowledge-guided lexical models and demonstrate their ability to measure individualism and collectivism. Our method (1) is highly scalable, (2) encodes domain knowledge from cultural psychology, and (3) does not require additional labeled data. 
    \item We validate our method against past collectivism research at the US state-level and present novel results at the US county level.
    \item We provide new insights into cultural variation across the US via a taxonomy of \textit{communities} (socio-demographic clusters of counties) from the American Communities Project~\cite{chinni2011our}.
    \item We highlight the failure of modern LLMs (GPT-3.5, GPT-4) to measure cultural variation or generate language that matches real-world cultural variation.
\end{enumerate}

\section{Building Knowledge-Guided Lexica}

\begin{figure*}[ht]
    \centering
    \includegraphics[width=\textwidth]{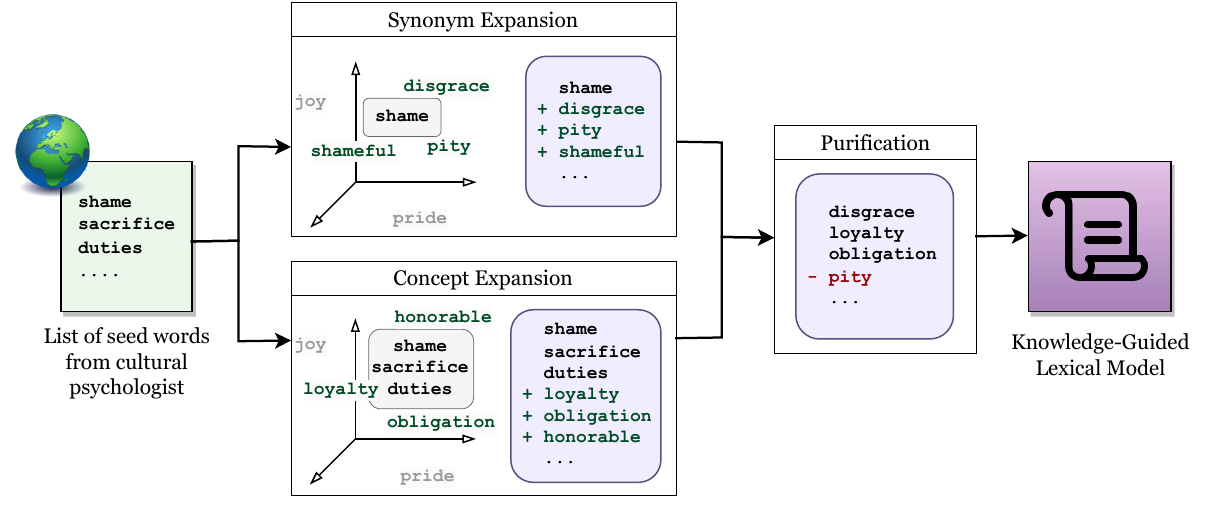}
    \caption{Our knowledge-guided lexica creation method. We begin with a set of seed words curated by an expert psychologist. The first stage, \textit{Expansion}, consists of synonym expansion and concept expansion, done in parallel. The second stage, \textit{Purification}, includes frequency-based and correlation-based pruning, done sequentially.}
    \label{fig:lexica-creation}
\end{figure*}

\paragraph{Issues with traditional lexica.} Lexica, or sets of curated words, are a highly scalable and explainable method for analyzing large datasets. However, building a full lexicon linked to cultural theory from the ground up is a time-intensive process, and sometimes takes psychologists years to complete \cite{pennebaker2015development}. Additionally, psychologists often make the error of including words in lexica that correlate negatively with the construct they are trying to measure. \citet{jaidka2020estimating} find that removing certain words from lexical categories in LIWC 2015 \cite{pennebaker2015development} actually improves overall performance. 

These erroneous words can be caused by the following two phenomena:

\begin{enumerate}
    \vspace{-0.1cm}
    \itemsep 0pt
    \item \textit{Inflated Frequency:} Highly frequent words in a lexical category can correlate negatively with their counterparts (e.g. words like ``love'' and ``lol'', contained in the LabMT positive sentiment lexicon \cite{dodds2011temporal} correlate negatively with happiness \cite{jaidka2020estimating}) and muddy the results of a lexicon.
    \item \textit{Polysemy}: Words like ``tender,'' which describe positive emotion, may have other meanings (e.g. describing chicken/steak, or referring to financial tenders), thus reducing the effectiveness of a lexicon.
\end{enumerate}

To mitigate this, we propose a method that still builds on domain knowledge, but produces a lexicon that is \textit{internally coherent}, bypassing the issues introduced by inflated frequency and polysemy. 

Our approach has two components: Expansion and Purification. Figure~\ref{fig:lexica-creation} details this approach for the collectivism lexicon we generate. Using this method, we utilize domain knowledge from an expert psychologist to create a lexical model that has wider coverage of our corpus. We can then use this model to analyze our geolocated Twitter corpus and measure regional variation in individualism and collectivism across the US.

\begin{table*}[t]
    \small
    \begin{tabular}{lp{11.5cm}}  
        \toprule
        \textbf{Collectivism Seed Words} &  \texttt{duties, responsibilities, role, fit in, community, sacrifice, shame, required, rules, honor, support, rely, loyal, respect, obedience} \\
        \midrule
        \textbf{Individualism Seed Words} &  \texttt {humans, humanity, worldwide, universal, mankind, everyone, collective, global, equity, imagination, cooperate, cooperation, shared, joint, identity, guilt, diversity}\\
        \bottomrule
    \end{tabular}
    \caption{Seed words hypothesized to identify individualism and collectivism on social media, provided by an expert cultural psychologist. These words provide interesting insights into these two constructs --- according to our domain expert, ``collective'' and ``cooperation'' are actually individualist words. This is because collectivism emphasizes close relationships over strangers, whereas individualism emphasizes strangers and weak ties, hence the usage of words like ``collective''.}
    \label{tab:seed}
\end{table*}

\paragraph{Step 1: Seed Word Generation} We first ask an expert psychologist who has researched individualism and collectivism for many years to generate two small sets of seed words that capture each of these constructs (see Table~\ref{tab:seed}). However, a small set of seed words may not be enough to sufficiently analyze a corpus of 1.5$+$ billion Tweets and may have erroneous words as described above. 

\paragraph{Step 2: Expansion} 
Next, we utilize word embeddings\footnote{We use FastText~\cite{bojanowski2017enriching} due to its fixed vocabulary size, efficient nearest neighbors functionality, and ability to find synonyms in context-free scenarios, but our methods are more general and agnostic to embedding type.} to expand the set of seed words in two ways: we locate words that are similar to each seed word (\textit{synonym expansion}), and locate the words that are similar to the overall construct described by the complete set of seed words (\textit{concept expansion}). 

For synonym expansion, we find the nearest neighbors for each seed word in embedding space and add these neighboring words to our lexica. For example, in Figure~\ref{fig:lexica-creation}, we expand the word ``shame'' and add ``disgrace'', ``shameful'', and ``pity'' to our collectivism lexicon. We use cosine similarity to determine the nearest neighbors.

For concept expansion, we average the embeddings of each seed word set to find the \textit{centroid embeddings}. For example, to find the collectivism centroid embedding, we would average the embeddings of ``shame'', ``sacrifice'', ``duties'', etc. Then we find the nearest neighbors of each centroid embedding. Concept expansion adds other words to a lexical category that are similar to the overall concept described by the seed word set. 

By using embedding space to expand our lexica, we can additionally calculate a \textit{weight} for each expanded word. The weight for each seed word is $1$, and the weight for each word added during expansion is the cosine similarity to the corresponding seed word or centroid embedding. Highly similar words have a weight closer to $1$, while more distant words have a lower weight. Our final lexicon is the union of seed words, words added via synonym expansion, and words added via concept expansion.

This method is highly tunable --- any embeddings can be used, and the number of nearest neighbors returned during expansion can be adjusted based on the desired length of the final lexicon. To control the length of our final lexicon, we set two thresholds in this process: one for synonym expansion and one for concept expansion. We explore the effect of these expansion thresholds in Section~\ref{sec:ablation}.

\paragraph{Step 3: Purification} Upon aggregating the words returned from both expansion types, we want to ensure that the resulting lexicon is both pertinent and internally correlated. Namely, we want to avoid the pitfalls of traditional lexica, where erroneous words may lower the overall performance.

To ensure pertinence, we filter out rare words, or any words below a given usage frequency \cite{bojanowski2017enriching}. 
Next, we ensure internal correlation. We apply our lexica to our US Twitter Corpus and compute the weighted frequencies for each word at the county-level. 

Equation~\ref{eq:weight-freq} details how we compute $F(w_i)$, the weighted frequency for a word $w$ in County $i$, where $T_i$ refers to the subset of Tweets geolocated in County $i$, and $\mathrm{count}(w, t)$ refers to the number of times word $w$ appears in Tweet $t$.

\vspace{-0.2cm}
\begin{equation}
\label{eq:weight-freq}
    F(w_i) = \sum_{t  \in  T_i} \mathrm{weight}_w * \mathrm{count}(w, t) 
\end{equation}

\noindent To avoid issues that arise with inflated frequency and polysemy, we want to ensure that there are no words that correlate negatively with the other words in the lexicon. Specifically, we ensure that the product-moment correlation detailed in Equation~\ref{eq:corr} is greater than some positive threshold for every word $w_i$ in our lexicon $L$.

\vspace{-0.2cm}
\begin{equation}
\label{eq:corr}
    r(F(w_i), \sum_{w \in L \setminus w_i} F(w))
\end{equation}

\noindent If a word does not meet this criteria, we remove it from the lexicon. This purification step ensures that every word contributes correctly to measuring the relevant cultural dimension. We explore the effect of this purification threshold in Section~\ref{sec:ablation}.

Figure~\ref{fig:wordclouds} visualizes our expanded and purified knowledge-guided individualism and collectivism lexica. Note that this method can be used to measure regional variation for any cultural construct\footnote{Other example cultural constructs include power distance \cite{hofstede2011dimensionalizing} or looseness/tightness \cite{gelfand2006nature} and are also hypothesized to vary regionally.} by changing the seed words accordingly.

\begin{table*}[t]
    \centering
    \small
    \begin{tabular}{lrrrr@{\hskip 1cm}r}   
    \toprule
     & \textbf{\makecell[r]{Vandello \& Cohen's\\Collectivism Scores}} & \makecell[r]{\textbf{Grandparents} \\ \textbf{\textit{(GCI)}}} & \makecell[r]{\textbf{Religiosity} \\ \textit{\textbf{(GCI)}}} & \makecell[r]{\textbf{Ingroup Bias}\\ \textit{\textbf{(GCI)}}} & \makecell[r]{\textbf{Average}\\ \textbf{Validity}}\\ 
    \midrule
    \multicolumn{1}{l}{\textit{Collectivism ($\uparrow$ is better)}} \\
    \midrule
    \textbf{KGL Score (ours)} & 0.380$^{*}$ & 0.362$^{*}$ & 0.410$^{*}$ & 0.467$^{*}$ & \textbf{0.405} \\ [0.15cm]  
    \textbf{Seed Words Only} & -0.033 & -0.267 & -0.352$^*$ & -0.142 & -0.198 \\ [0.15cm] 
    \textbf{GPT-3.5 Baseline} & -0.094 & -0.094 & 0.056 & -0.035 & -0.042 \\[0.15cm]
    \midrule
     \multicolumn{1}{l}{\textit{Individualism ($\downarrow$ is better)}} \\
    \midrule
    \textbf{KGL Score (ours)} & \underline{-0.379$^{*}$} & \underline{-0.571$^*$} & \underline{-0.659$^*$} & \underline{-0.515$^*$} & \textbf{-0.531} \\[0.15cm]   
    \textbf{Seed Words Only} & \underline{-0.509$^*$} & \underline{-0.423$^*$} & \underline{-0.564$^*$} & \underline{-0.525$^*$} & -0.506 \\[0.15cm]   
    \textbf{GPT-3.5 Baseline} & -0.346$^*$ & -0.222 & 0.058 & 0.058 & -0.113 \\[0.15cm]
    \bottomrule
    \end{tabular}
    \caption{Pairwise product-moment correlations between our knowledge-guided lexica (KGL) scores and collectivism validation variables. We use Vandello \& Cohen's Collectivism Scores and collectivism indicators from the Global Collectivism Index (GCI) at the US-state level. $^*$ indicates correlation is significant ($p < 0.05$). Underlined correlations are not found to be significantly different after bootstrapping test. We see that our method (KGL) outperforms both baselines --- using only the seed words provided by an expert psychologist, and zero-shot labeling a subset of Tweets from each state using GPT-3.5.}
    \label{tab:validation}
\end{table*}

\section{Evaluation}
Upon expanding and purifying the lexica, we apply it to our Twitter Corpus. To get a collectivism score for county $C$, we sum the weighted frequencies of each word $w$ in our collectivism lexicon $L_{\mathrm{coll}}$, as outlined in Equation~\ref{eq:score}.

\vspace{-0.2cm}
\begin{equation}
\label{eq:score}
    \mathrm{Collectivism}(C) = \sum_{w  \in  L_\mathrm{coll}} F(w) 
\end{equation}

\noindent We then aggregate these county-level scores to the state-level, and validate our results using past state-level collectivism research from \citet{vandello1999patterns, pelham2022truly}. We expect our state-level collectivism results to correlate positively with these indicators. As there are no similar resources for individualism, we use the strength of negative correlation to evaluate our state-level individualism results. 

\paragraph{Vandello \& Cohen's Scores.} \citet{vandello1999patterns} conduct a survey-based study of individualism and collectivism in the US and rank all states from most to least collectivist. We use these rankings as our first validation variable. 

\paragraph{GCI Indicators.} We also use relevant collectivism indicators from the Global Collectivism Index \cite{pelham2022truly} using corresponding questions from the 2017 US census and the 2017 wave of the World Values Survey \cite{haerpfer2020world}.

All six variables in the Global Collectivism Index – total fertility rate, living arrangements (\% households with people over 60 and children under 14), stability of marriage (divorce rate to marriage rate ratio), religiosity, collective transportation, and ingroup bias (approximated by compatriotism due to lack of state-level data) – are replicable at the state-level using US census data and WVS data. Note that when aggregating US census data from county-level to state-level, we weight each county equally, due to disproportionate amounts of data coming from big cities. 

In order to determine which of these six replicated variables also measures collectivism within the United States, we sample subsets of the six variables and use Cronbach’s alpha to maximize internal consistency. We limit the subsets to size three or larger, following \citet{pelham2022truly}'s validation of three collectivism indicators per nation. Upon exploring all possible subsets of size three or more, we find that the set of living arrangements, religiosity, and ingroup bias yields the highest Cronbach’s alpha (0.702); we choose these as our additional validation variables. 

\paragraph{Income.} As a sanity check, we additionally validate against median household income at the state-level. Prior research has found that higher income levels lead to more individualistic values \cite{pelham2022truly}. Similarly, we find that median household income correlates positively with our individualism lexicon scores (0.431) and negatively with our collectivism lexicon scores (-0.288).

\begin{figure*}[t]
    \centering
    \includegraphics[width=0.9\textwidth]{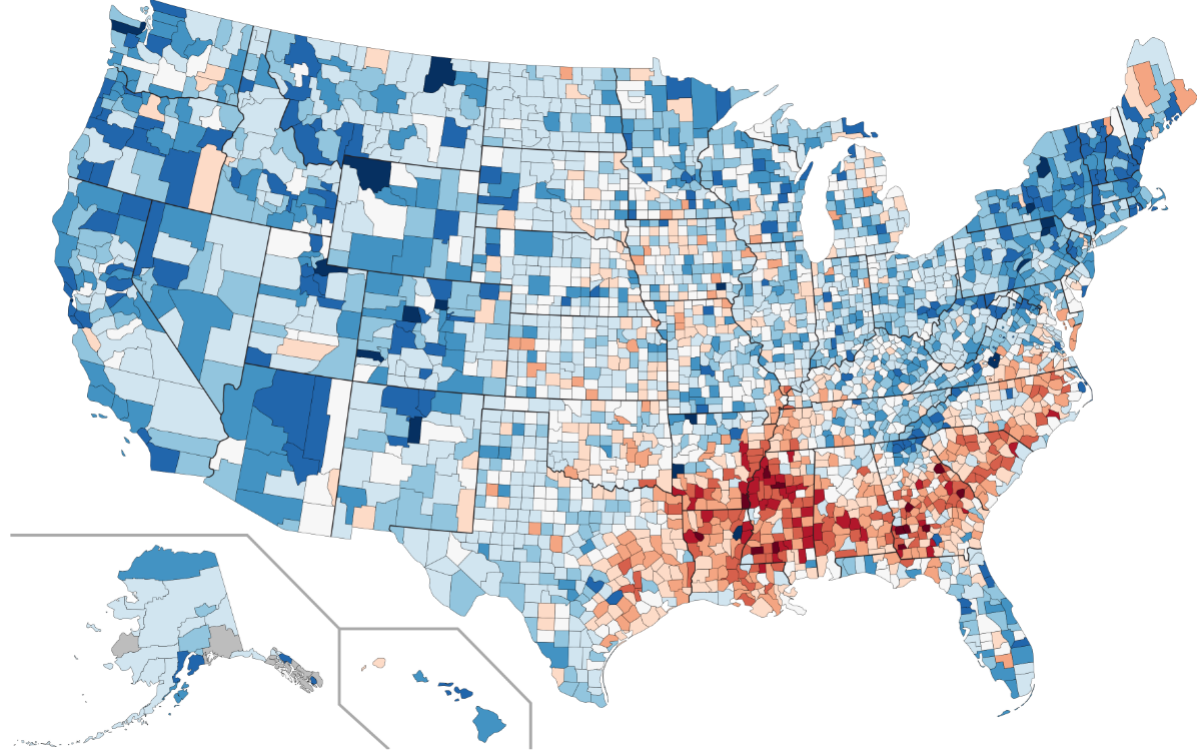}
    \caption{Collectivism (red) and individualism (blue) across US counties. Dark red = higher collectivism and dark blue = higher individualism. We include 2042 counties with sufficient data to compute individualism/collectivism scores, along with 1095 counties with interpolated scores based on geographic and socio-demographic variables.}
    \label{fig:map}
\end{figure*}

\section{Results}

\begin{figure*}[t]
    \centering
    \includegraphics[width=\textwidth]{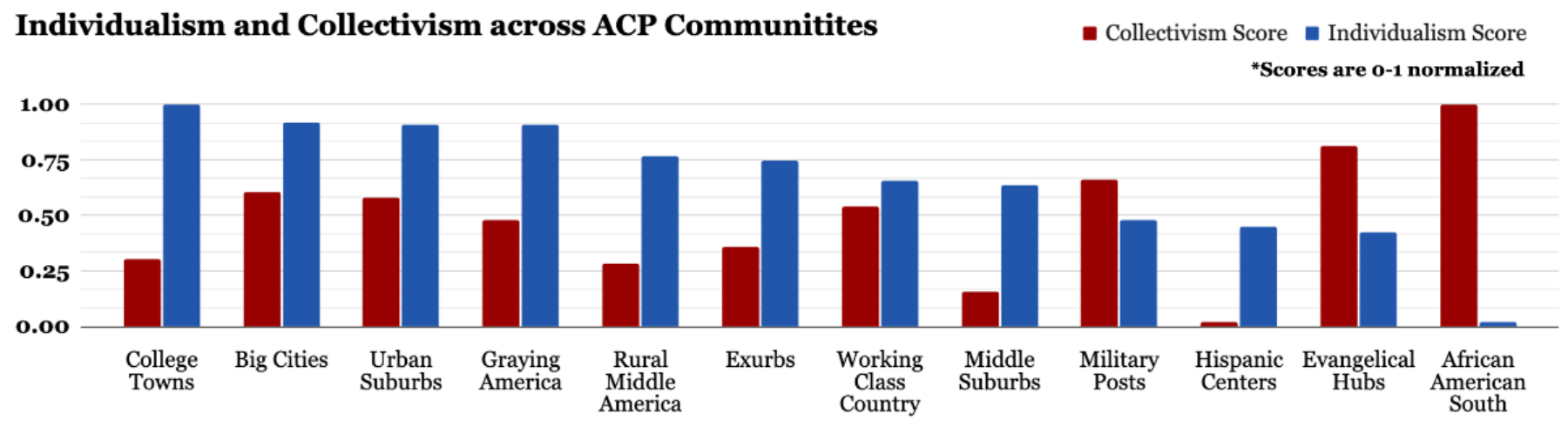}
    \caption{A comparison of collectivism (red) and individualism (blue) scores across communities defined by the American Communities Project, ordered from most individualistic (left) to least individualistic (right). We only analyze communities with over 40 included counties. \textit{Scores are 0-1 normalized.}
}
    \label{fig:graph}
\end{figure*}

\vspace{0.25cm}
\noindent Table~\ref{tab:validation} contains the correlations between our collectivism and individualism lexicon scores and each of the four validation variables, across US states. We observe that collectivist word use positively correlates with all validation outcomes, and individualist word use correlates negatively. We also observe a strong negative correlation (-0.510) between our individualism and collectivism scores at the US state level. 

To further assess the success of our method, we compare our correlations against the following baselines:
\begin{itemize}
    \vspace{-0.2cm}
    \itemsep 0cm
    \item \textbf{Seed Words Only:} To analyze the efficacy of expansion $+$ purification, we compare against solely using the expert-curated seed words.
    \item \textbf{GPT-3.5 Baseline:} To explore whether our method outperforms prompting a pre-trained LM, we subsample a total of 100,000 Tweets (2,000 per state) from our corpus and have GPT-3.5 label each Tweet as individualist, collectivist, or neither. We then calculate $\frac{num(\mathrm{individualist})}{2,000}$ and $\frac{num(\mathrm{collectivist})}{2,000}$ as the corresponding individualism and collectivism scores for that state. See Appendix~\ref{app:gpt-baseline} for additional details on prompting procedure.
\end{itemize}

\paragraph{Interpreting correlations in Table~\ref{tab:validation}.} There is little past work measuring culture in NLP, so we rely on state-level measures of collectivism to validate against, as no county-level measures exist. The magnitude of the effect sizes in Table~\ref{tab:validation} align with previous work. \citealt{giorgi2021regional} use this Twitter data to estimate 5-factor personality across US states. The average correlations across all five personality dimensions between these estimates and several national personality surveys range between 0.29 and 0.61. 

Given our correlations fall within the range established in prior work, we conclude our lexica successfully capture individualism and collectivism.

\paragraph{Measuring county-level variation.} Upon confirming the validity of our lexica, we apply them to county-level geolocated Tweets, as detailed in Equation~\ref{eq:score}, to gain a more fine-grained understanding of how individualism and collectivism vary regionally.\footnote{We release our lexica, county-level and state-level scores, and relevant code at \url{https://github.com/shreyahavaldar/knowledge_guided_lexica}} Figure~\ref{fig:map} illustrates this variation, plotting the difference between the individualism and collectivism score. The deep south shows high levels of collectivism (dark red) and low levels of individualism (light blue). Conversely, the West Coast and the Northeast show low levels of collectivism (light red) and high levels of individualism (dark blue). 

\paragraph{County interpolation.} We interpolate individualism and collectivism across US counties where Twitter data is not available, in order to provide county-level estimates for the entire country. 

To do this, we build a Gaussian Process (GP) regression model, which is traditionally used for spatial interpolation (i.e., kriging; \citealp{cressie1990origins}). Instead of interpolating based on physical proximity alone, we follow \citealt{giorgi2023filling} and interpolate over both physical and socio-demographic space, training the GP model on latitude and longitude coordinates of the county centroids and 11 socio-demographic variables. See Appendix Section \ref{app:interpolations} for additional details.

\paragraph{Community-level insights.} Cultural similarity is not always based on geographical proximity; two cities hundreds of miles apart may be more similar than a city and a rural farm a few miles away \cite{GUNTUKU20214034}. To show how county-level analyses of culture can help us better understand \textit{communities}, we additionally use 15 community types (e.g., College Towns, Urban Suburbs) identified by the American Communities Project (ACP). The ACP identified these communities based on socio-demographic attributes, not spatial clusters of counties. Previous studies have used these community types to identify cultural variation in excessive alcohol consumption~\cite{giorgi2020cultural} and self-reported physical and mental health~\cite{aggarwal2023cross,mangalik2023robust}.

Figure~\ref{fig:graph} shows county-level individualism and collectivism scores grouped into their corresponding ACP community (see Table~\ref{tab:county-stats} for counts.) These results provide novel insights into how culture varies regionally. For example, College Towns and Big Cities are highly individualist. These areas are also more affluent and have higher rates of education \cite{americancommunities}. This fits with prior research findings that people who are wealthy or educated tend to be more individualistic \cite{binder2019redistribution}. 

In contrast, the data shows that Evangelical Hubs and the African-American South are highly collectivist. These communities are tight-knit and religious areas \cite{americancommunities}, which have been linked to collectivism \cite{pelham2022truly}. Military Posts are also more collectivist, which fits with the tight ties in military service and ``duty to one's troop.'' This insight is helpful because we know of no cultural psychology research comparing military communities with civilian communities. Overall, our community-level findings are in line with prior work, and we introduce novel measurements for understudied communities.

\section{Ablation Studies}
\label{sec:ablation}

\subsection{Effect of Expansion Thresholds}
To evaluate the effect of our expansion thresholds, we conduct a hyperparameter search with \{0.7, 0.75, 0.8\} as the search space for our synonym expansion threshold, and \{0.4, 0.45, 0.5\} as the sample space for our cluster expansion. We run our pipeline 9 times, testing each of these combinations. Table~\ref{tab:exp-ablation} contains the validation results for each of our hyperparameter runs. 

We find that expansion greatly helps the validity of our collectivism lexicon, as our seed words do not perform well --- a lower cosine similarity threshold yields higher validation scores. Interestingly, we observe the opposite phenomenon for our individualism seed words --- more expansion hurts validity scores. We select \{0.75, 0.45\} for our final synonym and concept thresholds, respectively, as they yield the highest average validity. 

As we see in the case of our collectivism lexicon, the expansion step is crucial to gathering the words that make our lexicon perform well. However, the results are unstable across thresholds. This is expected, as the resulting lexica are not internally coherent. Upon ensuring internal coherence, our lexica become significantly more stable. 

\subsection{Effect of Purification Threshold}
To evaluate the effect of our purification threshold, we conduct another hyperparameter search over the space \{0, 0.05, 0.1, 0.15, 0.2\}. Here, we find that the purification threshold has very little impact on the overall validation score. As long as we ensure all words internally correlate positively (i.e. Equation~\ref{eq:corr} is greater than 0), the resulting lexicon is stable. We select 0.15 as our final purification threshold as this yields the highest average validity. Table~\ref{tab:pur-ablation} contains the validation results for each purification threshold.

Some examples of words removed due to negative internal correlation include ``benefit'', ``supporting'', and ``community'' for collectivism, and ``harmony'', ``evolution'' and ``international'' for individualism. Though these words are added during expansion as synonyms of seed words, they have different usage patterns than their counterparts. Some are erroneous words due to polysemy --- ``benefit'' may indicate helpful behavior, but may also refer to a fundraiser or gala event. Likewise, ``harmony'' may indicate unity, but is more frequently used to refer to music and melodies.

The purification component of our pipeline results in lexica that are highly stable across thresholds as well as reliable, highlighting the importance of internal coherence.

\section{Investigating Cultural Variation in LLM-generated Text}

With a validated method to measure regional variation in individualism and collectivism, we can now answer the question: \textit{Given geographical information, can LLMs generate text that mimics real-world cultural variation?}

\paragraph{Generating state-specific text.} A recent line of NLP research investigates how LLMs express personality, and explores imbuing a fixed set of characteristics into an LLM to create a persona \cite{mao2023editing, safdari2023personality, jiang2024personallm}. Drawing from this line of work, we create a geographic persona for an LLM (i.e. specifying a US state of residency), thus allowing us to generate state-specific text.

Specifically, we aim to recreate our dataset using an LLM, so we can directly compare whether synthetic LLM-generated text reflects the cultural variation found in Tweets from real people. We select four states for this experiment -- New York, Massachusetts, Louisiana, and Mississippi, as they have the highest levels of individualism (NY, MA) and collectivism (LA, MS), while also containing Tweets from the vast majority of their counties. 

Next, we prompt GPT-3.5 to generate Tweets as Twitter users living in each state. Following \citet{jiang2024personallm}, we use a temperature of $0.7 $ to encourage creativity and variance when emulating different users. We keep our geographic persona prompt concise and open-ended, so as not to skew the LLM with prior notions of expected topics or writing style. We generate 100,000 total Tweets (500 users per state; 50 Tweets per user) and sample a parallel subset from our real-world Twitter dataset with an identical user breakdown.

\begin{figure}[t]
    \centering
\includegraphics[width=\linewidth]{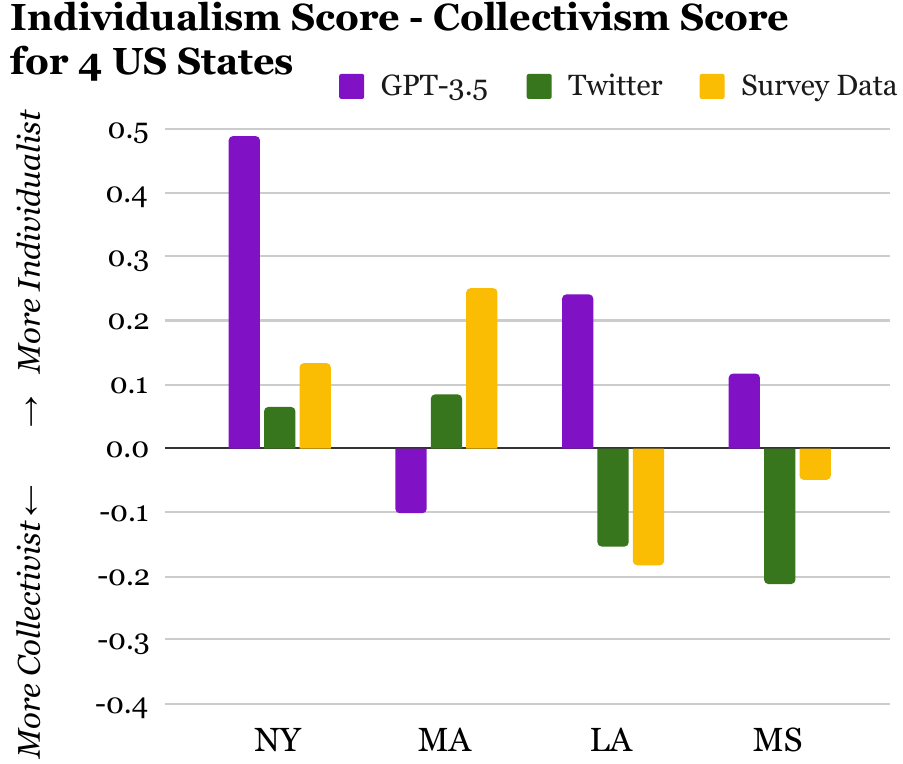}
    \caption{Individualism score minus collectivism score for LLM-generated and real-world Tweets. Across four US states, Twitter data (green) more closely aligns with Vandello \& Cohen's survey-based scores (yellow) compared to the GPT-3.5 data (purple). }
    \label{fig:gpt-generation}
\end{figure}

See Appendix~\ref{app:gpt-generation} for full details on experimental setup and prompting procedure. Sample Tweets generated by GPT-3.5 are shown below:

\vspace{0.25cm}
\small
\noindent \texttt{Nothing beats a slice of New York pizza on a Friday night \#NYCeats}

\vspace{0.25cm}

\noindent \texttt{Louisiana summers are no joke, y'all. The humidity is on another level. Stay cool out there, friends! \#LouisianaLife}

\normalsize

\paragraph{LLMs don't reflect cultural variation.} We then run our knowledge-guided lexical model on both datasets, and calculate the difference between the individualism and collectivism scores per state. The results are shown in Figure~\ref{fig:gpt-generation}, along with Vandello \& Cohen's collectivism scores\footnote{As Vandello \& Cohen's scores range from $0-1$, we shift the range by subtracting them from $0.5$, thus ensuring a higher number indicates more individualism.}. 

The Twitter scores match what we expect: NY and MA are more individualist, while LA and MS are more collectivist. However, the GPT scores do not reflect this pattern. Upon inspecting the generated Tweets, we notice a deeper problem -- the Tweets predominantly focus on state stereotypes (e.g. NY Tweets referencing pizza and bagels, LA Tweets referencing crawfish and Mardi Gras, etc.) 

In fact, this phenomenon causes New York to have a very high individualism score, due to frequently appearing phrases like ``the greatest city in the world'' and ``diversity of cultures.'' However, the unstable and incorrect results for the other three states indicate that GPT-3.5 cannot reliably mimic real-world cultural variation. Rather, the generated Tweets are highly stereotypical and cover a small fraction of topics found in Tweets from real people.

\section{Related Work}

\paragraph{Lexicon Induction.} Developing lexica to measure psychological and social constructs is a computationally inexpensive venture that can provide results at par with sophisticated LLMs. Emotion lexica \cite{mohammad2010emotions}, Demographic lexica \cite{sap2014developing}, and Politeness lexica \cite{hayati2021does, havaldar2023comparing} are a few such examples. Similar to prior work, we begin with an expert-curated list of seed words that we expand using semantic knowledge in LLMs. 

\paragraph{Measuring Cultural Constructs.}  
Individualism and collectivism have previously been studied to understand a range of constructs \cite{hamilton-etal-2016-inducing, hofstede2011dimensionalizing, triandis1993collectivism}. 

One of the earliest attempts to measure collectivism in the US uses a questionnaire-based survey approach \cite{vandello1999patterns}. More recently, \citet{bazzi2020frontier} uses infrequent names (common names indicate the desire to fit in vs. stand out \cite{twenge2010fitting}) as an indicator of individualism in their county-level study. \citet{chen2021culture} use citizen ancestry data to account for immigrants' culture. However, static approaches including name and ancestry mapping ignore the transient nature of the collectivism-individualism dimension and the fact that it evolves with the socio-economic environment \cite{lomas2023complexifying, santos2017global}. Question-based surveys may provide a truer picture but can become increasingly expensive and time-consuming when collecting granular data (e.g. county-level). In this scenario, social media language can help in dynamically modeling collectivism within regions \cite{aggarwal2023cross}.

\paragraph{Culture and NLP.} Detecting culture in LLMs and the consequent goal of "cultural alignment" are emerging research problems that often rely on decade-old measures of cultural constructs derived from a small sample of the population \cite{kovavc2023large,jin2023cultural,masoud2023cultural}. Social psychology has great potential in adapting machine-generated text for cross-cultural interactions \cite{marmolejo2023social}. Through this paper, we propose a highly scalable approach to dynamically measure cultural constructs at high granularity from publicly available social media language.

\section{Conclusion \& Future Work}

We introduce a new problem for the NLP community: \textit{measuring regional variation in culture}. This is a new class of problem --- obtaining labels for culture across states or countries is often infeasible, as there are very few labeled data points to train on. Using pre-trained LLMs to label data is unreliable, as these models lack basic cultural awareness. Additionally, scaling to label billions of data points is computationally infeasible. Classic deep learning methods fail to solve this problem; therefore, measuring culture necessitates a different approach.

We present a method to efficiently measure cultural variation by leveraging domain knowledge from cultural psychology to create knowledge-guided lexica. Our lexica filter out erroneous words and ensure internal coherence, bypassing the pitfalls of traditional lexica. By applying these lexica to social media language, we can estimate cultural differences at fine-grained geographic levels, such as states, counties, and communities -- a task that modern LLMs fail to accomplish. 

Future work could build on this method to get deeper insights into communities and cultures. For example, our method could be used to identify Tweets that mark cultural differences; we encourage researchers to build more sophisticated models on these identified Tweets. Additionally, our method is easily extendable to other cultural dimensions, such as power distance, tightness/looseness, etc. This method could also measure cultural variation globally, which requires analyzing different languages. Since our method is language agnostic, it can easily extend to non-English settings by leveraging multilingual embeddings.

\section{Limitations}

While we label each county for individualism and collectivism, we note that regions do not have a single culture. Within all regions, there is heterogeneity of cultural values and beliefs.  Since we use an open-source Twitter corpus, we also have poor coverage of counties with little to no Twitter data. We also only use Twitter users to represent each region, which may lead to an incomplete representation of these regions. Additionally, not all aspects of culture are revealed in language -- we are limited to analyzing only what people say online.

For our seed words, we only consult one domain expert. As a result, our final lexica are based on this expert's interpretation of individualism and collectivism. Extensions of this work could consult multiple experts to ensure downstream lexica are as unbiased as possible. 

Furthermore, our results are influenced by the choice of embedding model used for expansion. Additional work is needed to determine the effect of the embedding model, namely, whether a different model yields drastically different results. We also risk propagating any bias present in the embedding model during expansion.

In our analyses, we do not control for race, income, or other demographic variables. We know cultural values are correlated with some demographic variables --- for example, collectivism and individualism vary with income. Future work can improve upon these estimates by accounting for individual demographics. Additionally, it is unclear if this method of measuring cultural variation will work for all cultural dimensions. For example, power distance \cite{hofstede2011dimensionalizing} involves the relationship dynamics of two people, which might make it difficult to capture with lexica.

\section{Ethical Considerations}

The goal of studying cultural variation is to better understand cultures, not individuals. Nonetheless, the characterization of culture has the danger of stereotyping individuals. Individuals within each culture vary greatly. Studying culture can help us understand differences in psychology, but we should not assume that a cultural average will definitely apply to a particular individual from that culture.

All data used in this study is publicly available. While geolocated Twitter data is used, only aggregated spatial-level data is reported. That is, no person-level identifiable information is used or released for this study. 

\bibliography{references}

\begin{thebibliography}{51}
\expandafter\ifx\csname natexlab\endcsname\relax\def\natexlab#1{#1}\fi

\bibitem[{ACP(2023)}]{americancommunities}
ACP. 2023.
\newblock Home - american communities project - americancommunities.org.
\newblock \url{https://www.americancommunities.org/}.
\newblock [Accessed 14-08-2023].

\bibitem[{Aggarwal et~al.(2023)Aggarwal, Rai, Giorgi, Havaldar, Sherman,
  Mittal, and Guntuku}]{aggarwal2023cross}
Arnav Aggarwal, Sunny Rai, Salvatore Giorgi, Shreya Havaldar, Garrick Sherman,
  Juhi Mittal, and Sharath~Chandra Guntuku. 2023.
\newblock A cross-modal study of pain across communities in the united states.
\newblock In \emph{Companion Proceedings of the ACM Web Conference 2023}, pages
  1050--1058.

\bibitem[{Araque et~al.(2020)Araque, Gatti, and
  Kalimeri}]{araque2020moralstrength}
Oscar Araque, Lorenzo Gatti, and Kyriaki Kalimeri. 2020.
\newblock Moralstrength: Exploiting a moral lexicon and embedding similarity
  for moral foundations prediction.
\newblock \emph{Knowledge-based systems}, 191:105184.

\bibitem[{Arora et~al.(2023)Arora, Kaffee, and
  Augenstein}]{arora-etal-2023-probing}
Arnav Arora, Lucie-aim{\'e}e Kaffee, and Isabelle Augenstein. 2023.
\newblock \href {https://aclanthology.org/2023.c3nlp-1.12} {Probing pre-trained
  language models for cross-cultural differences in values}.
\newblock In \emph{Proceedings of the First Workshop on Cross-Cultural
  Considerations in NLP (C3NLP)}, pages 114--130, Dubrovnik, Croatia.
  Association for Computational Linguistics.

\bibitem[{Bazzi et~al.(2020)Bazzi, Fiszbein, and
  Gebresilasse}]{bazzi2020frontier}
Samuel Bazzi, Martin Fiszbein, and Mesay Gebresilasse. 2020.
\newblock Frontier culture: The roots and persistence of “rugged
  individualism” in the united states.
\newblock \emph{Econometrica}, 88(6):2329--2368.

\bibitem[{Binder(2019)}]{binder2019redistribution}
Carola~Conces Binder. 2019.
\newblock Redistribution and the individualism--collectivism dimension of
  culture.
\newblock \emph{Social Indicators Research}, 142(3):1175--1192.

\bibitem[{Bojanowski et~al.(2017)Bojanowski, Grave, Joulin, and
  Mikolov}]{bojanowski2017enriching}
Piotr Bojanowski, Edouard Grave, Armand Joulin, and Tomas Mikolov. 2017.
\newblock Enriching word vectors with subword information.
\newblock \emph{Transactions of the association for computational linguistics},
  5:135--146.

\bibitem[{Buechel et~al.(2020)Buechel, R{\"u}cker, and
  Hahn}]{buechel2020learning}
Sven Buechel, Susanna R{\"u}cker, and Udo Hahn. 2020.
\newblock Learning and evaluating emotion lexicons for 91 languages.
\newblock \emph{arXiv preprint arXiv:2005.05672}.

\bibitem[{Chen et~al.(2021)Chen, Frey, and Presidente}]{chen2021culture}
Chinchih Chen, Carl~Benedikt Frey, and Giorgio Presidente. 2021.
\newblock Culture and contagion: Individualism and compliance with covid-19
  policy.
\newblock \emph{Journal of economic behavior \& organization}, 190:191--200.

\bibitem[{Chinni and Gimpel(2011)}]{chinni2011our}
Dante Chinni and James Gimpel. 2011.
\newblock \emph{Our patchwork nation: The surprising truth about the" real"
  America}.
\newblock Penguin.

\bibitem[{Cohen(2001)}]{cohen2001cultural}
Dov Cohen. 2001.
\newblock Cultural variation: considerations and implications.
\newblock \emph{Psychological bulletin}, 127(4):451.

\bibitem[{Cressie(1990)}]{cressie1990origins}
Noel Cressie. 1990.
\newblock The origins of kriging.
\newblock \emph{Mathematical geology}, 22:239--252.

\bibitem[{Dodds et~al.(2011)Dodds, Harris, Kloumann, Bliss, and
  Danforth}]{dodds2011temporal}
Peter~Sheridan Dodds, Kameron~Decker Harris, Isabel~M Kloumann, Catherine~A
  Bliss, and Christopher~M Danforth. 2011.
\newblock Temporal patterns of happiness and information in a global social
  network: Hedonometrics and twitter.
\newblock \emph{PloS one}, 6(12):e26752.

\bibitem[{Gardner et~al.(2018)Gardner, Pleiss, Weinberger, Bindel, and
  Wilson}]{gardner2018gpytorch}
Jacob Gardner, Geoff Pleiss, Kilian~Q Weinberger, David Bindel, and Andrew~G
  Wilson. 2018.
\newblock Gpytorch: Blackbox matrix-matrix gaussian process inference with gpu
  acceleration.
\newblock \emph{Advances in neural information processing systems}, 31.

\bibitem[{Gelfand et~al.(2006)Gelfand, Nishii, and Raver}]{gelfand2006nature}
Michele~J Gelfand, Lisa~H Nishii, and Jana~L Raver. 2006.
\newblock On the nature and importance of cultural tightness-looseness.
\newblock \emph{Journal of applied psychology}, 91(6):1225.

\bibitem[{Geng et~al.(2022)Geng, Wu, Santhosh, Srivastava, Ungar, and
  Sedoc}]{geng2022inducing}
Yilin Geng, Zetian Wu, Roshan Santhosh, Tejas Srivastava, Lyle Ungar, and
  Jo{\~a}o Sedoc. 2022.
\newblock Inducing generalizable and interpretable lexica.
\newblock In \emph{Findings of the Association for Computational Linguistics:
  EMNLP 2022}, pages 4430--4448.

\bibitem[{Giorgi et~al.(2023)Giorgi, Eichstaedt, Preotiuc-Pietro, Gardner,
  {Andrew Schwartz}, and Ungar}]{giorgi2023filling}
Salvatore Giorgi, Johannes~C. Eichstaedt, Daniel Preotiuc-Pietro, Jacob~R.
  Gardner, H.~{Andrew Schwartz}, and Lyle~H. Ungar. 2023.
\newblock \href {https://doi.org/https://doi.org/10.1016/j.cresp.2023.100159}
  {Filling in the white space: Spatial interpolation with gaussian processes
  and social media data}.
\newblock \emph{Current Research in Ecological and Social Psychology}, page
  100159.

\bibitem[{Giorgi et~al.(2021)Giorgi, Nguyen, Eichstaedt, Kern, Yaden, Kosinski,
  Seligman, Ungar, Andrew~Schwartz, and Park}]{giorgi2021regional}
Salvatore Giorgi, Khoa~Le Nguyen, Johannes~C Eichstaedt, Margaret~L Kern,
  David~B Yaden, Michal Kosinski, Martin~EP Seligman, Lyle~H Ungar,
  H~Andrew~Schwartz, and Gregory Park. 2021.
\newblock Regional personality assessment through social media language.
\newblock \emph{Journal of Personality}.

\bibitem[{Giorgi et~al.(2018)Giorgi, Preo{\c{t}}iuc-Pietro, Buffone, Rieman,
  Ungar, and Schwartz}]{giorgi-etal-2018-remarkable}
Salvatore Giorgi, Daniel Preo{\c{t}}iuc-Pietro, Anneke Buffone, Daniel Rieman,
  Lyle Ungar, and H.~Andrew Schwartz. 2018.
\newblock \href {https://doi.org/10.18653/v1/D18-1148} {The remarkable benefit
  of user-level aggregation for lexical-based population-level predictions}.
\newblock In \emph{Proceedings of the 2018 Conference on Empirical Methods in
  Natural Language Processing}, pages 1167--1172, Brussels, Belgium.
  Association for Computational Linguistics.

\bibitem[{Giorgi et~al.(2020)Giorgi, Yaden, Eichstaedt, Ashford, Buffone,
  Schwartz, Ungar, and Curtis}]{giorgi2020cultural}
Salvatore Giorgi, David~B Yaden, Johannes~C Eichstaedt, Robert~D Ashford,
  Anneke~EK Buffone, H~Andrew Schwartz, Lyle~H Ungar, and Brenda Curtis. 2020.
\newblock Cultural differences in tweeting about drinking across the us.
\newblock \emph{International journal of environmental research and public
  health}, 17(4):1125.

\bibitem[{Guntuku et~al.(2021)Guntuku, Buttenheim, Sherman, and
  Merchant}]{GUNTUKU20214034}
Sharath~Chandra Guntuku, Alison~M. Buttenheim, Garrick Sherman, and Raina~M.
  Merchant. 2021.
\newblock \href {https://doi.org/https://doi.org/10.1016/j.vaccine.2021.06.014}
  {Twitter discourse reveals geographical and temporal variation in concerns
  about covid-19 vaccines in the united states}.
\newblock \emph{Vaccine}, 39(30):4034--4038.

\bibitem[{Haerpfer et~al.(2020)Haerpfer, Inglehart, Moreno, Welzel, Kizilova,
  Diez-Medrano, Lagos, Norris, Ponarin, Puranen et~al.}]{haerpfer2020world}
Christian Haerpfer, Ronald Inglehart, Alejandro Moreno, Christian Welzel,
  Kseniya Kizilova, Jaime Diez-Medrano, Marta Lagos, Pippa Norris, Eduard
  Ponarin, Bi~Puranen, et~al. 2020.
\newblock World values survey: Round seven--country-pooled datafile.
\newblock \emph{Madrid, Spain \& Vienna, Austria: JD Systems Institute \& WVSA
  Secretariat}, 7:2021.

\bibitem[{Hamilton et~al.(2016)Hamilton, Clark, Leskovec, and
  Jurafsky}]{hamilton-etal-2016-inducing}
William~L. Hamilton, Kevin Clark, Jure Leskovec, and Dan Jurafsky. 2016.
\newblock \href {https://doi.org/10.18653/v1/D16-1057} {Inducing
  domain-specific sentiment lexicons from unlabeled corpora}.
\newblock In \emph{Proceedings of the 2016 Conference on Empirical Methods in
  Natural Language Processing}, pages 595--605, Austin, Texas. Association for
  Computational Linguistics.

\bibitem[{Havaldar et~al.(2023{\natexlab{a}})Havaldar, Pressimone, Wong, and
  Ungar}]{havaldar2023comparing}
Shreya Havaldar, Matthew Pressimone, Eric Wong, and Lyle Ungar.
  2023{\natexlab{a}}.
\newblock \href {http://arxiv.org/abs/2310.07135} {Comparing styles across
  languages}.

\bibitem[{Havaldar et~al.(2023{\natexlab{b}})Havaldar, Rai, Singhal, Liu,
  Guntuku, and Ungar}]{havaldar-etal-2023-multilingual}
Shreya Havaldar, Sunny Rai, Bhumika Singhal, Langchen Liu, Sharath~Chandra
  Guntuku, and Lyle Ungar. 2023{\natexlab{b}}.
\newblock \href {https://aclanthology.org/2023.wassa-1.19} {Multilingual
  language models are not multicultural: A case study in emotion}.
\newblock In \emph{Proceedings of the 13th Workshop on Computational Approaches
  to Subjectivity, Sentiment, {\&} Social Media Analysis}, pages 202--214,
  Toronto, Canada. Association for Computational Linguistics.

\bibitem[{Hayati et~al.(2021)Hayati, Kang, and Ungar}]{hayati2021does}
Shirley~Anugrah Hayati, Dongyeop Kang, and Lyle Ungar. 2021.
\newblock Does bert learn as humans perceive? understanding linguistic styles
  through lexica.
\newblock In \emph{Proceedings of the 2021 Conference on Empirical Methods in
  Natural Language Processing}, pages 6323--6331.

\bibitem[{Hofstede(2011)}]{hofstede2011dimensionalizing}
Geert Hofstede. 2011.
\newblock Dimensionalizing cultures: The hofstede model in context.
\newblock \emph{Online readings in psychology and culture}, 2(1):8.

\bibitem[{Hovy and Yang(2021)}]{hovy-yang-2021-importance}
Dirk Hovy and Diyi Yang. 2021.
\newblock \href {https://doi.org/10.18653/v1/2021.naacl-main.49} {The
  importance of modeling social factors of language: Theory and practice}.
\newblock In \emph{Proceedings of the 2021 Conference of the North American
  Chapter of the Association for Computational Linguistics: Human Language
  Technologies}, pages 588--602, Online. Association for Computational
  Linguistics.

\bibitem[{Jaidka et~al.(2020)Jaidka, Giorgi, Schwartz, Kern, Ungar, and
  Eichstaedt}]{jaidka2020estimating}
Kokil Jaidka, Salvatore Giorgi, H~Andrew Schwartz, Margaret~L Kern, Lyle~H
  Ungar, and Johannes~C Eichstaedt. 2020.
\newblock Estimating geographic subjective well-being from twitter: A
  comparison of dictionary and data-driven language methods.
\newblock \emph{Proceedings of the National Academy of Sciences},
  117(19):10165--10171.

\bibitem[{Jiang et~al.(2024)Jiang, Zhang, Cao, Breazeal, Kabbara, and
  Roy}]{jiang2024personallm}
Hang Jiang, Xiajie Zhang, Xubo Cao, Cynthia Breazeal, Jad Kabbara, and Deb Roy.
  2024.
\newblock \href {http://arxiv.org/abs/2305.02547} {Personallm: Investigating
  the ability of large language models to express personality traits}.

\bibitem[{Jin et~al.(2023)Jin, Zhang, Shu, and Cui}]{jin2023cultural}
Chuanyang Jin, Songyang Zhang, Tianmin Shu, and Zhihan Cui. 2023.
\newblock The cultural psychology of large language models: Is chatgpt a
  holistic or analytic thinker?
\newblock \emph{arXiv preprint arXiv:2308.14242}.

\bibitem[{Kova{\v{c}} et~al.(2023)Kova{\v{c}}, Sawayama, Portelas, Colas,
  Dominey, and Oudeyer}]{kovavc2023large}
Grgur Kova{\v{c}}, Masataka Sawayama, R{\'e}my Portelas, C{\'e}dric Colas,
  Peter~Ford Dominey, and Pierre-Yves Oudeyer. 2023.
\newblock Large language models as superpositions of cultural perspectives.
\newblock \emph{arXiv preprint arXiv:2307.07870}.

\bibitem[{Liu et~al.(2023)Liu, Koto, Baldwin, and
  Gurevych}]{liu2023multilingual}
Chen~Cecilia Liu, Fajri Koto, Timothy Baldwin, and Iryna Gurevych. 2023.
\newblock \href {http://arxiv.org/abs/2309.08591} {Are multilingual llms
  culturally-diverse reasoners? an investigation into multicultural proverbs
  and sayings}.

\bibitem[{Lomas et~al.(2023)Lomas, Diego-Rosell, Shiba, Standridge, Lee, Case,
  Lai, and VanderWeele}]{lomas2023complexifying}
Tim Lomas, Pablo Diego-Rosell, Koichiro Shiba, Priscilla Standridge, Matthew~T
  Lee, Brendan Case, Alden~Yuanhong Lai, and Tyler~J VanderWeele. 2023.
\newblock Complexifying individualism versus collectivism and west versus east:
  Exploring global diversity in perspectives on self and other in the gallup
  world poll.
\newblock \emph{Journal of Cross-Cultural Psychology}, 54(1):61--89.

\bibitem[{Mangalik et~al.(2023)Mangalik, Eichstaedt, Giorgi, Mun, Ahmed, Gill,
  Ganesan, Subrahmanya, Soni, Clouston et~al.}]{mangalik2023robust}
Siddharth Mangalik, Johannes~C Eichstaedt, Salvatore Giorgi, Jihu Mun, Farhan
  Ahmed, Gilvir Gill, Adithya~V Ganesan, Shashanka Subrahmanya, Nikita Soni,
  Sean~AP Clouston, et~al. 2023.
\newblock Robust language-based mental health assessments in time and space
  through social media.
\newblock \emph{arXiv preprint arXiv:2302.12952}.

\bibitem[{Mao et~al.(2023)Mao, Zhang, Wang, Wang, Yao, Jiang, Xie, Huang, and
  Chen}]{mao2023editing}
Shengyu Mao, Ningyu Zhang, Xiaohan Wang, Mengru Wang, Yunzhi Yao, Yong Jiang,
  Pengjun Xie, Fei Huang, and Huajun Chen. 2023.
\newblock Editing personality for llms.
\newblock \emph{arXiv preprint arXiv:2310.02168}.

\bibitem[{Marmolejo-Ramos and Tejada(2023)}]{marmolejo2023social}
Fernando Marmolejo-Ramos and Julian Tejada. 2023.
\newblock Social psychology: Spotting social faux pas with ai.
\newblock \emph{Communications Psychology}, 1(1):17.

\bibitem[{Masoud et~al.(2023)Masoud, Liu, Ferianc, Treleaven, and
  Rodrigues}]{masoud2023cultural}
Reem~I Masoud, Ziquan Liu, Martin Ferianc, Philip Treleaven, and Miguel
  Rodrigues. 2023.
\newblock Cultural alignment in large language models: An explanatory analysis
  based on hofstede's cultural dimensions.
\newblock \emph{arXiv preprint arXiv:2309.12342}.

\bibitem[{Mohammad and Turney(2010)}]{mohammad2010emotions}
Saif Mohammad and Peter Turney. 2010.
\newblock Emotions evoked by common words and phrases: Using mechanical turk to
  create an emotion lexicon.
\newblock In \emph{Proceedings of the NAACL HLT 2010 workshop on computational
  approaches to analysis and generation of emotion in text}, pages 26--34.

\bibitem[{Oishi et~al.(2009)Oishi, Diener, Lucas, and Suh}]{oishi2009cross}
Shigehiro Oishi, Ed~Diener, Richard~E Lucas, and Eunkook~M Suh. 2009.
\newblock Cross-cultural variations in predictors of life satisfaction:
  Perspectives from needs and values.
\newblock \emph{Culture and well-being: The collected works of Ed Diener},
  pages 109--127.

\bibitem[{Pelham et~al.(2022)Pelham, Hardin, Murray, Shimizu, and
  Vandello}]{pelham2022truly}
Brett Pelham, Curtis Hardin, Damian Murray, Mitsuru Shimizu, and Joseph
  Vandello. 2022.
\newblock A truly global, non-weird examination of collectivism: The global
  collectivism index (gci).
\newblock \emph{Current Research in Ecological and Social Psychology},
  3:100030.

\bibitem[{Pennebaker et~al.(2015)Pennebaker, Boyd, Jordan, and
  Blackburn}]{pennebaker2015development}
James~W Pennebaker, Ryan~L Boyd, Kayla Jordan, and Kate Blackburn. 2015.
\newblock The development and psychometric properties of liwc2015.
\newblock Technical report.

\bibitem[{Safdari et~al.(2023)Safdari, Serapio-Garc{\'\i}a, Crepy, Fitz,
  Romero, Sun, Abdulhai, Faust, and Matari{\'c}}]{safdari2023personality}
Mustafa Safdari, Greg Serapio-Garc{\'\i}a, Cl{\'e}ment Crepy, Stephen Fitz,
  Peter Romero, Luning Sun, Marwa Abdulhai, Aleksandra Faust, and Maja
  Matari{\'c}. 2023.
\newblock Personality traits in large language models.
\newblock \emph{arXiv preprint arXiv:2307.00184}.

\bibitem[{Santos et~al.(2017)Santos, Varnum, and Grossmann}]{santos2017global}
Henri~C Santos, Michael~EW Varnum, and Igor Grossmann. 2017.
\newblock Global increases in individualism.
\newblock \emph{Psychological science}, 28(9):1228--1239.

\bibitem[{Sap et~al.(2014)Sap, Park, Eichstaedt, Kern, Stillwell, Kosinski,
  Ungar, and Schwartz}]{sap2014developing}
Maarten Sap, Gregory Park, Johannes Eichstaedt, Margaret Kern, David Stillwell,
  Michal Kosinski, Lyle Ungar, and H~Andrew Schwartz. 2014.
\newblock Developing age and gender predictive lexica over social media.
\newblock In \emph{Proceedings of the 2014 conference on empirical methods in
  natural language processing (EMNLP)}, pages 1146--1151.

\bibitem[{Sorensen et~al.(2023)Sorensen, Jiang, Hwang, Levine, Pyatkin, West,
  Dziri, Lu, Rao, Bhagavatula, Sap, Tasioulas, and Choi}]{sorensen2023value}
Taylor Sorensen, Liwei Jiang, Jena Hwang, Sydney Levine, Valentina Pyatkin,
  Peter West, Nouha Dziri, Ximing Lu, Kavel Rao, Chandra Bhagavatula, Maarten
  Sap, John Tasioulas, and Yejin Choi. 2023.
\newblock \href {http://arxiv.org/abs/2309.00779} {Value kaleidoscope: Engaging
  ai with pluralistic human values, rights, and duties}.

\bibitem[{Talhelm et~al.(2014)Talhelm, Zhang, Oishi, Shimin, Duan, Lan, and
  Kitayama}]{talhelm2014large}
Thomas Talhelm, Xuemin Zhang, Shigehiro Oishi, Chen Shimin, Dongyuan Duan,
  Xuezhao Lan, and Shinobu Kitayama. 2014.
\newblock Large-scale psychological differences within china explained by rice
  versus wheat agriculture.
\newblock \emph{Science}, 344(6184):603--608.

\bibitem[{Triandis(1993)}]{triandis1993collectivism}
Harry~C Triandis. 1993.
\newblock Collectivism and individualism as cultural syndromes.
\newblock \emph{Cross-cultural research}, 27(3-4):155--180.

\bibitem[{Tsai et~al.(2006)Tsai, Knutson, and Fung}]{tsai2006cultural}
Jeanne~L Tsai, Brian Knutson, and Helene~H Fung. 2006.
\newblock Cultural variation in affect valuation.
\newblock \emph{Journal of personality and social psychology}, 90(2):288.

\bibitem[{Twenge et~al.(2010)Twenge, Abebe, and Campbell}]{twenge2010fitting}
Jean~M Twenge, Emodish~M Abebe, and W~Keith Campbell. 2010.
\newblock Fitting in or standing out: Trends in american parents' choices for
  children’s names, 1880--2007.
\newblock \emph{Social Psychological and Personality Science}, 1(1):19--25.

\bibitem[{Vandello and Cohen(1999)}]{vandello1999patterns}
Joseph~A Vandello and Dov Cohen. 1999.
\newblock Patterns of individualism and collectivism across the united states.
\newblock \emph{Journal of personality and social psychology}, 77(2):279.

\end{thebibliography}
\bibliographystyle{acl_natbib}

\newpage

\appendix

\setcounter{table}{0}
\renewcommand{\thetable}{A\arabic{table}}

\section{Open-Source Twitter Corpus}
\label{app:corpus}
We use the County Tweet Lexical Bank, an open source data set of features extracted from a corpus of 1.5 billion tweets from approximately 6 million US county-mapped users~\cite{giorgi-etal-2018-remarkable}. While the full details of the dataset can be found in the original paper, we give a high-level summary to aid the reader. The dataset is built from a larger corpus which is a 10\% sample of Twitter from 2009-2015 (over 30 billion tweets). These tweets are then mapped to US counties via latitude and longitude coordinates associated with the tweets or self-reported location information in the Twitter user's profile (a free text field). A Twitter user is included in this data set if they have posted at least 30 or more English tweets, and a county is included if at least 100 such users are mapped to that respective county. This process resulted in 1.5 billion tweets mapped to over 2,000 US counties.

\section{Scalability Calculations}
\label{app:calculations}

We outline the proposed costs of using various LM-based techniques to label our corpus of 1.5 billion Tweets:

\paragraph{Proposed cost of GPT-4} As of August 2023, the OpenAI API rate for GPT-4 is \$0.06 cents per 1,000 tokens. Assuming 10 tokens per Tweet, we get:
\begin{equation}
    1.5e9 \: \mathrm{Tweets} \times \frac{10 \: \mathrm{Tokens}}{\mathrm{Tweet}} \times \frac{\$0.06}{1,000 \: \mathrm{Tokens}} 
\end{equation}
\noindent This yields a total cost of \$900,000.

\paragraph{Proposed cost of GPT-3.5} As of August 2023, the OpenAI API rate for GPT-3.5 is \$0.002 per 1,000 tokens. Assuming 10 tokens per Tweet, we get:
\begin{equation}
    1.5e9 \: \mathrm{Tweets} \times \frac{10 \: \mathrm{Tokens}}{\mathrm{Tweet}} \times \frac{\$0.002}{1,000 \: \mathrm{Tokens}} 
\end{equation}
\noindent This yields a total cost of \$30,000.

\begin{table}
    \centering
    \small
    \begin{tabular}{lr}
    \toprule
        \textbf{ACP Community} & \textbf{Num Counties} \\
        \midrule    
        Exurbs & 207 \\
        Graying America & 164 \\
        African American South & 252 \\
        Evangelical Hubs & 269 \\
        Working Class Country & 159 \\
        Military Posts & 70 \\
        Urban Suburbs & 103 \\
        College Towns & 151 \\
        Big Cities & 46 \\
        Hispanic Centers & 87 \\
        Rural Middle America & 403 \\
        Middle Suburbs & 77 \\
    \bottomrule
    \end{tabular}
    \caption{Number of included counties for each ACP community included in the analysis in Figure~\ref{fig:graph}.}
    \label{tab:county-stats}
\end{table}

\begin{figure}[t]
    \centering
    \includegraphics{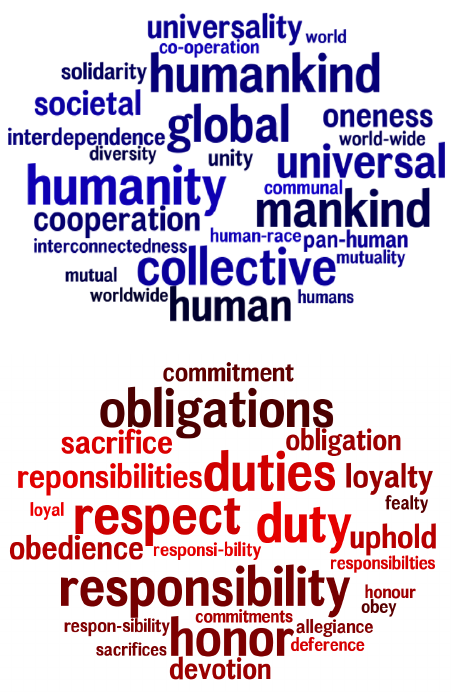}
    \caption{Word clouds visualizing our individualism lexica (blue, top) and collectivism lexica (red, bottom). Larger words have a higher weight, while smaller words have a lower weight.}
    \label{fig:wordclouds}
\end{figure}

\section{GPT-3.5 Baseline}
\label{app:gpt-baseline}
To assess whether a pre-trained LLM is capable of measuring individualism and collectivism, we sample a subset of our Twitter corpus (2,000 Tweets per state) and have GPT-3.5 assign a label to each Tweet. Our prompt is as follows: \\

\noindent \small \texttt{\textbf{System Prompt:} Given a Tweet, try to reason about whether it reflects the cultural dimension of Individualism or Collectivism. Collectivists are closely linked individuals who view themselves primarily as parts of a whole, be it a family, a network of co-workers, a tribe, or a nation. Such people are mainly motivated by the norms and duties imposed by the collective entity. Individualists are motivated by their own preferences, needs, and rights, giving priority to personal rather than group goals.} \\

\noindent \texttt{If the Tweet does not reflect either cultural dimension, please label it 'Neither'.} \\

\noindent \texttt{Example input and output: \\ Tweet: \{Tweet Text\}} \\
\noindent \texttt{Label: \{Individualism, Collectivism, Neither\}} \\

\noindent \small \texttt{\textbf{User Prompt:} Tweet: \{Tweet Text\}} \\

\normalsize

We provide GPT-3.5 with the definition of individualism and collectivism as defined by \citet{triandis1993collectivism}, and then have it label each tweet with one of three labels -- individualism, collectivism, or neither. Across all states, GPT-3.5 labels at least 50\% of the Tweets, ensuring enough signal for adequate comparison. We take cues from \citet{sorensen2023value}, which prompts LLMs for values, to design this prompt.

\section{Interpolations}
\label{app:interpolations}

We use 11 socio-demographic variables to interpolate individualism and collectivism across US counties with insufficient Twitter data. This includes four socioeconomic variables (median household income, percentage of the population with a Bachelor's degree, unemployment rate, and high school graduation rate) and seven demographic variables: (population density, median age, and the percentage of the population in rural areas, Hispanic, female, married, and African American). This is implemented with the GPyTorch package~\cite{gardner2018gpytorch} with a learning rate of 0.1 and 500 iterations over the training data. 

\begin{figure*}[t]
    \centering
    \includegraphics[width=0.9\textwidth]{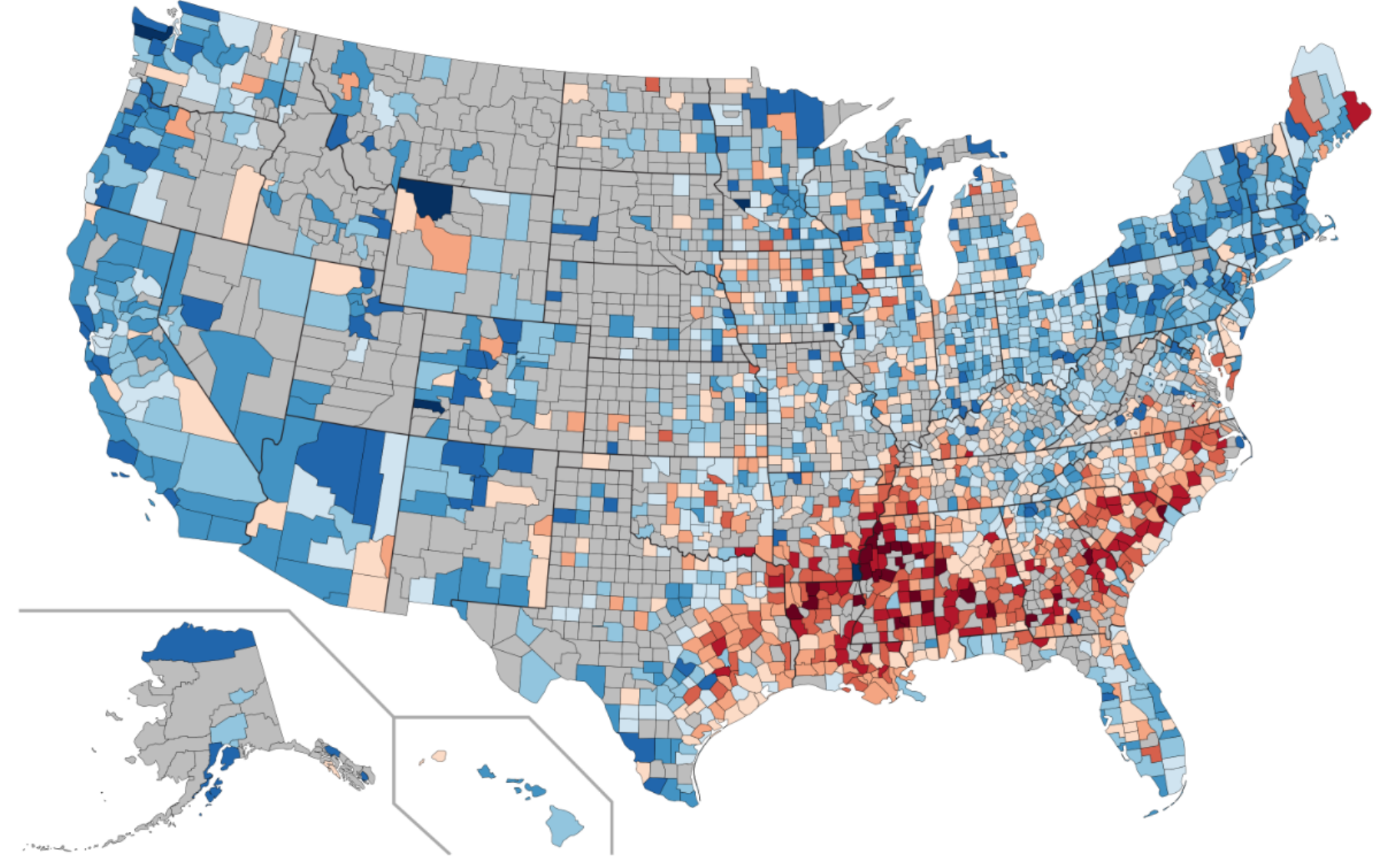}
    \caption{Collectivism (red) and individualism (blue) across US counties. Dark red = higher collectivism and dark blue = higher individualism. We show only the 2042 counties with sufficient data to compute individualism/collectivism scores. Gray counties do not have enough Twitter data to estimate scores.}
    \label{fig:map_original}
\end{figure*}

Figure~\ref{fig:map_original} shows the original 2042 counties, and Figure~\ref{fig:map} contains the interpolated counties as well.

\begin{table*}
\centering
    \small
    \begin{tabular}{rrrrrrrr}
    \toprule
         \textbf{\makecell[r]{Cluster\\Expansion\\Threshold}} & \textbf{\makecell[r]{Synonym\\Expansion\\Threshold}} & \textbf{\makecell[r]{Lexicon\\Length}} & \textbf{\makecell[r]{V\&C's\\Collectivism\\Scores}} & \makecell[r]{\textbf{Grandparents} \\ \textbf{\textit{(GCI)}}} & \makecell[r]{\textbf{Religiosity} \\ \textit{\textbf{(GCI)}}} & \makecell[r]{\textbf{Ingroup Bias}\\ \textit{\textbf{(GCI)}}} & \makecell[r]{\textbf{Average}\\ \textbf{Validity}}\\         
        \midrule
        \multicolumn{3}{l}{\textit{Collectivism ($\uparrow$ is better)}} \\
        \midrule
         0.4 & 0.7 & 366 & 0.273 & 0.504$^*$ & 0.531$^*$ & 0.552$^*$ & 0.465 \\
         0.4 & 0.75 & 366 & 0.273 & 0.504$^*$ & 0.531$^*$ & 0.552$^*$ & 0.465 \\
         0.4 & 0.8 & 366 & 0.273 & 0.504$^*$ & 0.531$^*$ & 0.552$^*$ & 0.465 \\
         \midrule
         0.45 & 0.7 & 100 & 0.226 & 0.113 & 0.124 & 0.274 & 0.184 \\
         0.45 & 0.75 & 98 & 0.275 & 0.171 & 0.175 & 0.314$^*$ & 0.234 \\
         0.45 & 0.8 & 95 & 0.263 & 0.168 & 0.167 & 0.305$^*$ & 0.226 \\
         \midrule
         0.5 & 0.7 & 42 & -0.014 & -0.234 & -0.297 & -0.094 & -0.16 \\
         0.5 & 0.75 & 39 & 0.029 & -0.184 & -0.26 & -0.057 & -0.118 \\
         0.5 & 0.8 & 34 & 0.008 & -0.196 & -0.279 & -0.078 & -0.137 \\
        \midrule
        \multicolumn{3}{l}{\textit{Individualism ($\downarrow$ is better)}} \\
        \midrule       
         0.4 & 0.7 & 479 & 0.095 & 0.512$^*$ & 0.446$^*$ & 0.308$^*$ & 0.34 \\
         0.4 & 0.75 & 479 & 0.095 & 0.512$^*$ & 0.446$^*$ & 0.308$^*$ & 0.34 \\
         0.4 & 0.8 & 479 & 0.095 & 0.512$^*$ & 0.446$^*$ & 0.308$^*$ & 0.34 \\
         \midrule
         0.45 & 0.7 & 119 & -0.417$^*$ & -0.545$^*$ & -0.664$^*$ & -0.542$^*$ & -0.542 \\
         0.45 & 0.75 & 119 & -0.417$^*$ & -0.545$^*$ & -0.664$^*$ & -0.542$^*$ & -0.542 \\
         0.45 & 0.8 & 119 & -0.417$^*$ & -0.545$^*$ & -0.664$^*$ & -0.542$^*$ & -0.542 \\
         \midrule
         0.5 & 0.7 & 37 & -0.545$^*$ & -0.504$^*$ & -0.618$^*$ & -0.539$^*$ & -0.552 \\
         0.5 & 0.75 & 37 & -0.545$^*$ & -0.504$^*$ & -0.618$^*$ & -0.539$^*$ & -0.552 \\
         0.5 & 0.8 & 37 & -0.545$^*$ & -0.504$^*$ & -0.618$^*$ & -0.539$^*$ & -0.552 \\
    \bottomrule
    \end{tabular}
    \caption{Ablation study investigating the effect of the expansion thresholds.}
    \label{tab:exp-ablation}
\end{table*}

\begin{table*}
    \centering
    \small
    \begin{tabular}{rrrrrrr}
    \toprule
         \textbf{\makecell[r]{Purification\\Threshold}} & \textbf{\makecell[r]{Lexicon\\Length}} & \textbf{\makecell[r]{V\&C's\\Collectivism\\Scores}} & \makecell[r]{\textbf{Grandparents} \\ \textbf{\textit{(GCI)}}} & \makecell[r]{\textbf{Religiosity} \\ \textit{\textbf{(GCI)}}} & \makecell[r]{\textbf{Ingroup Bias}\\ \textit{\textbf{(GCI)}}} & \makecell[r]{\textbf{Average}\\ \textbf{Validity}}\\         
        \midrule
        \multicolumn{3}{l}{\textit{Collectivism ($\uparrow$ is better)}} \\
        \midrule

        0 & 62 & 0.385$^*$ & 0.346$^*$ & 0.397$^*$ & 0.467$^*$ & 0.399 \\
        0.05 & 56 & 0.391$^*$ & 0.351$^*$ & 0.404$^*$ & 0.47$^*$ & 0.404 \\
        0.1 & 43 & 0.388$^*$ & 0.345$^*$ & 0.4$^*$ & 0.464$^*$ & 0.399 \\
        0.15 & 30 & 0.38$^*$ & 0.362$^*$ & 0.41$^*$ & 0.467$^*$ & 0.405 \\
        0.2 & 18 & 0.365$^*$ & 0.357$^*$ & 0.412$^*$ & 0.468$^*$ & 0.4 \\
        \midrule
        \multicolumn{3}{l}{\textit{Individualism ($\downarrow$ is better)}} \\
        \midrule  

        0 & 59 & -0.35$^*$ & -0.554$^*$ & -0.656$^*$ & -0.512$^*$ & -0.518 \\
        0.05 & 56 & -0.351$^*$ & -0.55$^*$ & -0.656$^*$ & -0.511$^*$ & -0.517 \\
        0.1 & 48 & -0.374$^*$ & -0.568$^*$ & -0.658$^*$ & -0.513$^*$ & -0.528 \\
        0.15 & 42 & -0.379$^*$ & -0.571$^*$ & -0.659$^*$ & -0.515$^*$ & -0.531 \\
        0.2 & 36 & -0.382$^*$ & -0.569$^*$ & -0.66$^*$ & -0.52$^*$ & -0.533 \\
    \bottomrule
    \end{tabular}    
    \caption{Ablation study investigating the effect of the purification threshold.}
    \label{tab:pur-ablation}
\end{table*}

\section{GPT-3.5 Tweet Generation}
\label{app:gpt-generation}
To assess whether a pre-trained LLM can generate language that reflects real-world variation in individualism and collectivism, we have GPT-3.5 generate 100,000 Tweets from four states - New York, Massachusetts, Louisiana, and Mississippi. 

\paragraph{Generating geographic personas.} When constructing our prompt, we instruct the LLM to behave like an individual living in a given state. We specify solely the state of residency, keeping the prompt concise and open-ended, so as not to bias the LLM towards a specific topic or style when generating text. 

\paragraph{Emulating users.} To construct a dataset parallel to our Twitter dataset, we generate Tweets in batches, with each query to GPT-3.5 emulating a single ``user''. To encourage creativity and variance in responses, we set the temperature relatively high ($0.7$), following \citet{jiang2024personallm}. Each query emulates a different ``user,'' with 50 Tweets generated in response to a single query. 

Our prompt is as follows: \\

\noindent \small \texttt{\textbf{System Prompt:} You are an individual who lives in \{STATE\}. You enjoy writing Tweets and posting them on Twitter for your followers to read.} \\

\noindent \texttt{\textbf{User Prompt:} Generate 50 Tweets that you might write and post online. Remember that each Tweet has a 280 character limit. Separate each Tweet with a newline character.} \\

\normalsize

We ensure that the resulting response meets the following criteria: (1) The response contains 50 separate Tweets, and (2) Each Tweet contains no more than 280 characters. All generated Tweets can be found at \url{https://github.com/shreyahavaldar/knowledge_guided_lexica}.

\end{document}